\documentclass[conference]{IEEEtran}
\IEEEoverridecommandlockouts
\usepackage{graphicx}
\usepackage{amsfonts}
\usepackage{amsmath}
\usepackage{cite}
\usepackage[dvipsnames]{xcolor}
\usepackage[ruled]{algorithm2e}
\usepackage{enumitem}
\usepackage{hyperref}
\hypersetup{
    colorlinks=true,
    linkcolor=blue,
    filecolor=magenta,      
    urlcolor=cyan,
    citecolor = {blue}
    }

\makeatletter
\def\ps@IEEEtitlepagestyle{%
  \def\@oddfoot{\mycopyrightnotice}%
  \def\@evenfoot{}%
}
\def\mycopyrightnotice{%
  {\footnotesize DISTRIBUTION A. Approved for public release; distribution unlimited. OPSEC \# 3733 \hfill}% <--- Change here
  \gdef\mycopyrightnotice{}% just in case
}

\begin{document}
\title{An Active Learning Framework for Constructing High-fidelity Mobility Maps}
\author{\IEEEauthorblockN{Gary R. Marple\IEEEauthorrefmark{1}, David Gorsich\IEEEauthorrefmark{2}, Paramsothy Jayakumar\IEEEauthorrefmark{2}, Shravan Veerapaneni\IEEEauthorrefmark{1}}
\IEEEauthorblockA{\IEEEauthorrefmark{1}Department of Mathematics, University of Michigan}
\IEEEauthorblockA{\IEEEauthorrefmark{2}U.S. Army CCDC Ground Vehicle Systems Center}
}
\maketitle

\noindent
\begin{abstract}
A mobility map, which provides maximum achievable speed on a given terrain, is essential for path planning of autonomous ground vehicles in off-road settings. While physics-based simulations play a central role in creating next-generation, high-fidelity mobility maps, they are cumbersome and expensive. For instance, a typical simulation can take weeks to run on a supercomputer and each map requires thousands of such simulations. Recent work at the U.S. Army CCDC Ground Vehicle Systems Center has shown that trained machine learning classifiers can greatly improve the efficiency of this process. However, deciding which simulations to run in order to train the classifier efficiently is still an open problem. According to PAC learning theory, data that can be separated by a classifier is expected to require $\mathcal{O}(1/\epsilon)$ randomly selected points (simulations) to train the classifier with error less than $\epsilon$. In this paper, building on existing algorithms, we introduce an active learning paradigm that substantially reduces the number of simulations needed to train a machine learning classifier without sacrificing accuracy. Experimental results suggest that our sampling algorithm can train a neural network, with higher accuracy, using less than half the number of simulations when compared to random sampling.
\end{abstract}

\section{Introduction}
\label{Introduction}
\noindent

Mobility is the essential requirement for all military ground vehicles and the loss of mobility due to unfavorable terrain or soil conditions can jeopardize a mission's success and can leave troops stranded. To avoid this scenario, a planner must use a mobility map that gives the maximum predicted speed the vehicles would be expected to reach while traversing a treacherous off-road terrain. The NATO Reference Mobility Model (NRMM) is widely used for predicting the mobility of ground vehicles \cite{haley1979nato, ahlvin1992nato}. However, newer vehicles containing advanced technologies have mobility capabilities that cannot easily be predicted using the NRMM method because of its empirical nature. As a result, the NATO Next Generation NRMM Team has identified physics-based modeling, mainly using the discrete element method (DEM), as being a potential high-fidelity method for predicting mobility \cite{dasch}.

\begin{figure*}
\begin{center}
\includegraphics[width=\textwidth]{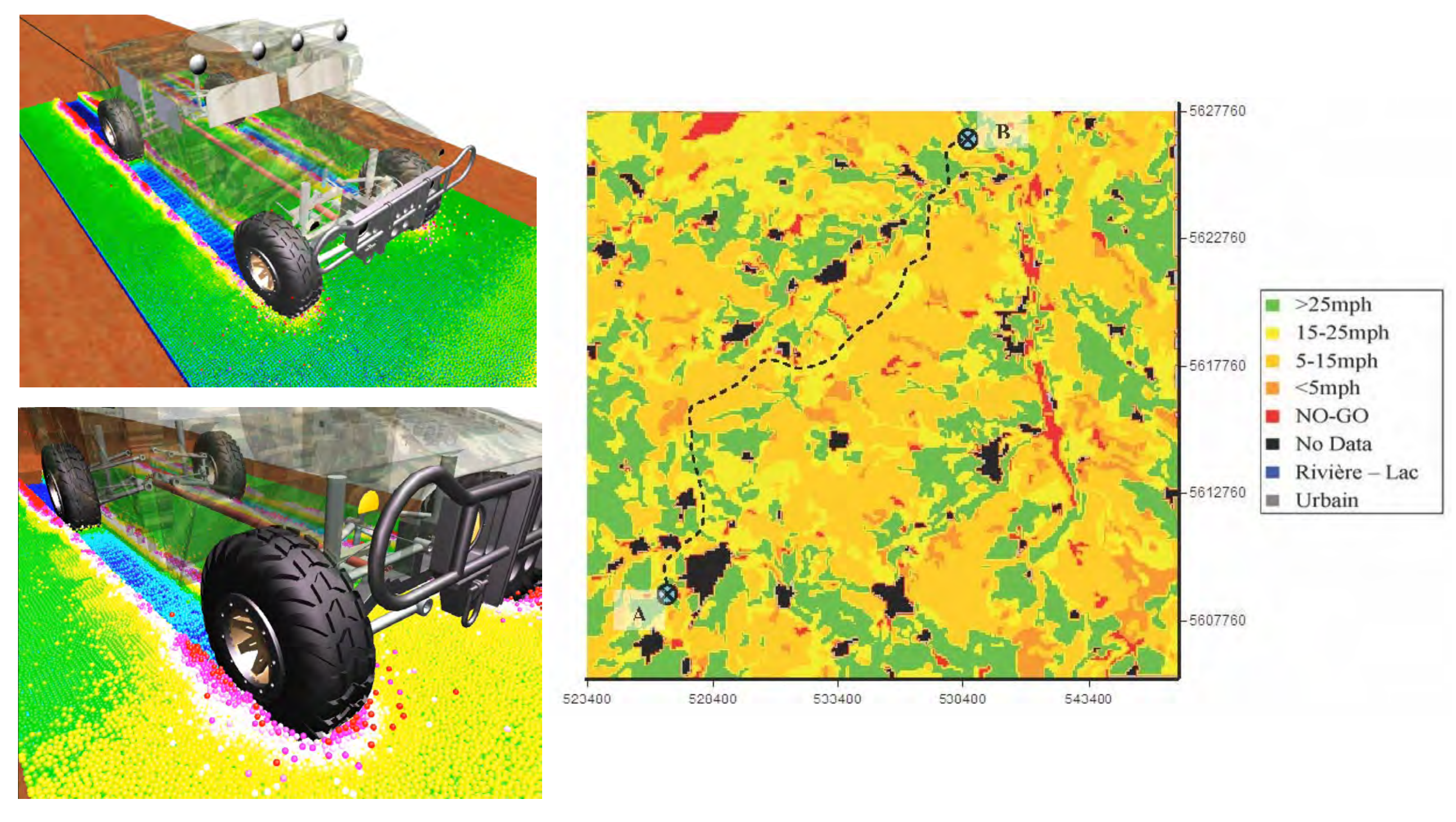}
\end{center}
\caption{(a) A DEM-based simulation of a vehicle traversing off-road conditions. The vehicle shown in this figure will be used for all experiments. (b) An example of a mobility map. The colors indicate the maximum sustained speed an off-road vehicle would be expected to reach.}
\label{SimMap}
\end{figure*}

While physics-based simulations can offer more accurate predictions of vehicle mobility, generating a mobility map that can accurately predict a mobility metric---such as the {\em speed-made-good}, defined as the ratio of the Euclidean distance and the time required to travel between two points, regardless of the actual path taken \cite{haley1979nato}---requires tens of thousands of simulations. Consequently, it can take weeks to generate such maps using high performance computing (HPC) architectures. Recent work by Mechergui and Jayakumar \cite{jayakumar} showed that machine learning classifiers can be trained to quickly generate mobility maps. While this approach is promising, it comes with its own challenges.\\ \indent
The main obstacle with using supervised learning is the computational expense that is incurred when constructing the training set. A computationally-intensive DEM simulation must be performed to predict the speed-made-good to label a single data point. This becomes prohibitively expensive since performing each numerical simulation can take over a week on a 20-core compute node using state-of-the-art simulation software; see \cite{wasfy2019understanding} for more details. Based on the usage cost of the HPC cluster utilized for running the numerical simulations in this paper (the Flux supercomputer at the University of Michigan \cite{flux}), {\em generating even a small training set with 200 points costs over ten thousand dollars.}
An additional issue is that simple sampling techniques, such as uniform random sampling, often choose uninformative data points that have little to no effect on the accuracy of the classifier. In other words, running simulations with parameter values that the classifier is already confident about only reinforces the model and does little to illuminate inaccuracies that can be improved upon. Finally, the focus of \cite{jayakumar} was on training classifiers with 2-dimensional feature spaces only, when in actuality, the speed-made-good depends on many parameters, such as the terrain topology and profile, soil type (mud, snow, sand, etc.), vegetation, and weather conditions. This poses a significant challenge due to the curse of dimensionality and the computational limitations that prevent us from generating large training sets.\\ \indent
Of course, some of these challenges may be alleviated by more efficient DEM simulations, algorithms for which are an active area of research \cite{negrut2012solving, corona2019tensor, yamashita2019hierarchical, de2019scalable}. However, even a moderate reduction in simulation times would not eliminate the need for machine learning-based predictions. Once trained, a machine learning classifier, such as a multilayer perceptron (MLP), can generate mobility maps on-the-fly. This is critically important since changing weather conditions can quickly alter soil properties. In addition, uncertainty quantification techniques that account for imprecise measurements can be significantly hindered if the model evaluation time is long. Machine learning classifiers offer a faster and more economical means of addressing these problems. However, training these algorithms in a reasonable time with an affordable budget is still a challenge.\\ \indent
To reduce the number of simulations, in this work, we developed an active learning-based approach that allows us to generate mobility maps using less data. Unlike supervised learning where the entire training set is constructed {\em a priori}, active learning allows the classifier to interactively query an annotator about a pool of unlabeled data \cite{settles}. Once the queried instances have been labeled, they are added to the training set, the classifier is retrained, and a new set of instances are chosen. In many cases, this iterative approach can train a classifier with higher accuracy using fewer instances when compared to supervised learning. Active learning has been utilized in many areas, including natural language processing, drug discovery, text classification, image retrieval, medical image classification, and landslide prediction \cite{thompson,reker,tong_doller,tong_edward,hoi, stumpf}. In general, an active learning-based approach is often advantageous when the application has a large amount of unlabeled data available, but labeling that data is expensive or time-consuming. This is precisely the situation that arises when training a classifier to generate mobility maps. In our case, the annotator is a computationally demanding DEM-based simulation, and the unlabeled pool that the active learner picks from consists of all possible combinations of soil parameters. Building on existing active learning algorithms, we propose a framework that is tailored to the needs of mobility map construction. \\ \indent
The paper is organized as follows. We will begin by giving a primer on active learning in Section \ref{Background}, where we review the query-by-bagging algorithm and discuss some of its advantages over uncertainty sampling. In Section \ref{MapGeneration}, we will focus on some of the main considerations that went into the development of our active learning approach tailored to mobility map generation. Section \ref{Experiments} will focus on our experimental results, where we used physics-based simulations to train a neural network to predict the speed-made-good using 2- and 3-dimensional feature spaces. Finally, Section \ref{Conclusions} will give our conclusions along with areas for future research.

\section{Preliminaries}
\label{Background}
\noindent
In this section, we begin by introducing the notion of a version space and show how it can be used to identify informative instances. This will lead to a discussion about the query-by-committee and query-by-bagging algorithms and why query-by-bagging is well-suited for predicting simulation results.

\subsection{Version Space}
\noindent
To begin with, assume that all instances have noise-free labels. To introduce the version space, we first need to define a hypothesis. A hypothesis $h:X\to Y$ is a function, generated by a machine learning algorithm, that maps instances $x$ in the feature space $X$ to labels $y$ in the set of class labels $Y$. A hypothesis space $\mathcal{H}$ is the set of all hypotheses under consideration. For example, a single set of values for the weights in a MLP could be used to specify a single hypothesis, and the set of all possible sets of values for the weights in a MLP could be thought of as being representative of a hypothesis space. The set of all hypotheses $h\in \mathcal{H}$ that are consistent with the training set $\mathcal{L}$ is referred to as the version space $\mathcal{V}$ \cite{mitchell}. To be precise,
$$\mathcal{V}=\left\{h\in \mathcal{H} | h(x)=y\text{ for all }\left<x,y\right>\in \mathcal{L} \right\}.$$
\indent
Now suppose that $\mathcal{V}$ is bounded and that there exists a hypothesis $c\in\mathcal{H}$ that can correctly label any instance in $X$. This hypothesis would be consistent with any training set, so we can conclude that $c\in \mathcal{V}$. We would like to query instances that allow us to focus in on $c$ by reducing the size of $\mathcal{V}$. If we choose to query an instance whose label happens to be consistent with all of the hypotheses in $\mathcal{V}$, then $\mathcal{V}$ would remain unchanged after the instance was added to $\mathcal{L}$. Clearly, this is something we would like to avoid. On the other hand, if we query an instance that has a label that is inconsistent with some of the hypotheses in $\mathcal{V}$, then those hypotheses would be eliminated from $\mathcal{V}$. Since the label for a point is unknown in advance, a good approach is to query instances where any label will result in the elimination of a large portion of $\mathcal{V}$. In other words, we should try to query points that create the greatest amount of disagreement.  For example, we could try to query instances where at least half of the hypotheses disagree, no matter the label. This would cut the size of $\mathcal{V}$ by at least a half each time we queried a point, giving us exponential convergence onto $c$. The main challenge with doing this is in determining when the majority of the hypotheses in $\mathcal{V}$ disagree. Fortunately, the query-by-committee algorithm gives us a way of approximating this.

\subsection{Query-By-Committee}

\noindent
The query-by-committee (QBC) algorithm is a highly effective active learning technique that focuses on reducing the version space by forming a committee of hypotheses that serve as a representative sample of $\mathcal{V}$ \cite{seung}. This committee of hypotheses determines whether a point should be queried or not by following a voting process. If the majority of the hypotheses in the committee disagree on the label for a point, then querying that point will eliminate at least half of the committee members from $\mathcal{V}$. Furthermore, if the committee is representative of $\mathcal{V}$, then we would expect that $\mathcal{V}$ would be approximately halved as well. Under certain conditions, QBC has been shown to achieve prediction error $\epsilon$ with high probability using $\mathcal{O}(1/\epsilon)$ unlabeled instances and $\mathcal{O}\left(\log(1/\epsilon)\right)$ queries \cite{freund}, that is, an exponential improvement over random sampling.

While such a promising theoretical guarantee is encouraging, it turns out that QBC has few practical applications. Part of the reason for this is that noisy labels can lead to scenarios where $\mathcal{V}$ is empty. In other words, noisy labels can result in the elimination of hypotheses that would otherwise be consistent with the data. As a result, this can make it impossible to form a committee. An additional issue can also occur when $\mathcal{V}$ is nonempty. When using a deterministic classifier, such as a SVM, it can be difficult to find multiple hypotheses that are consistent with the data \cite{campbell}. The answer to both of these issues is to randomize the training set using a technique called query-by-bagging \cite{abe}.

\begin{algorithm}[t]
\caption{Query-By-Committee}
{\bf Input:} Number of trials: $N$\\
Number of committee members: $n_c$\\
A randomized classifier: $A$\\
{\bf Initialize:} $\mathcal{L}_1$, $\mathcal{U}_1$ with random instances\\
\For{$k=1,\dots,N$}{
\begin{enumerate}[label=\arabic*., leftmargin=0.3cm,rightmargin=0.5cm]
\item Train $A$ on $\mathcal{L}_k$ and generate $h_1,\dots,h_{n_c}$.
\item Choose a point $x^*\in \mathcal{U}_k$ that satisfies $\max_{y\in Y}\left| \left\{ t\leq n_c|h_t(x)=y\right\}\right|\leq n_c/2$.\\
\item Query $x^*$ to obtain $y^*=\text{Oracle}(x^*)$.
\item $\mathcal{L}_{k+1}=\mathcal{L}_k\cup\left\{\left<x^*,y^*\right>\right\},\quad \mathcal{U}_{k+1}=\mathcal{U}_k\setminus \{x^*\}$\\
\end{enumerate}
}
{\bf Output:} $h(x)=\arg\max_{y\in Y}\left| \left\{ t\leq n_c|h_t(x)=y\right\}\right|,$\\
where $h_t$ are hypotheses of the $N$th stage.
\end{algorithm}

\subsection{Query-By-Bagging}
\noindent
The query-by-bagging (QBag) algorithm, first introduced by Abe and Mamitsuka \cite{abe}, can be thought of as a combination of QBC and bagging \cite{breiman}. Like QBC, the method uses an ensemble of classifiers to make querying decisions. However, instead of training all committee members on the same data, QBag trains each classifier on a subset of the training set. Usually, these subsets are constructed by randomly sampling (with replacement) from the original training data. This allows deterministic classifiers, such as a SVM, to generate multiple predictions while still being trained with data that has a similar distribution to the initial training set. Like QBC, instances are queried if there is a significant amount of disagreement among the committee members.

\subsection{Comparison with Uncertainty Sampling}
\noindent
Uncertainty sampling is a popular approach for performing active learning \cite{lewis}. It has been used with many different applications and has been shown to reduce the number of queried instances under certain conditions. In this work, we avoided uncertainty sampling; we would like to highlight some of the reasons why. Let's begin with a brief overview of the approach. The main idea behind uncertainty sampling is simple: query instances where the model is most uncertain about the label. There are numerous ways to do this. For example, an active learner may query the instance with the greatest Shannon entropy, which is given by
$$\phi_\text{ENT}(x)=-\sum_y{P_\theta(y|x)\log_2{P_\theta(y|x)}},$$
where $P_\theta(y|x)$ is the predicted probability, generated by the classifier, that an unlabeled instance $x$ will have label $y$ given a set of model parameters $\theta$. The result for this and many other metrics is that most of the queried instances tend to lie on or near the decision boundaries. While sampling along decision boundaries can help the classifier make more accurate predictions, there are some additional considerations that shouldn't be overlooked.

To begin with, uncertainty sampling bases its queries on the predictions made by a single classifier, which is often trained using limited data. As a result, the classifier may make highly erroneous assumptions about the location of the decision boundaries. Most likely, this would cause the learner to query uninformative points, which in our case, would result in a poor utilization of our computing resources. This problem is made worse when points are queried in batches, since each point in the batch is queried using the same inaccurate classifier. For our application, multiple simulations will be performed in parallel, so batch sampling is a must. An additional concern has to do with noisy data. While all learning algorithms tend to perform worse on noisy data, uncertainty sampling is especially vulnerable. This is due to the classifier relying on a small set of unreliable data, which can easily lead to inaccurate predictions.

With QBag, an ensemble of classifiers is used to make querying decisions. This means that a few inaccurate classifiers would not necessarily alter the querying results. Instead, the ensemble tends to have an averaging effect where highly erroneous predictions can be balanced out by the more sensible ones. An additional benefit is that each classifier in the ensemble is trained using a slightly different training set. This means that a mislabeled point has less influence on the querying process since some of the classifiers will be unaware of its existence. For additional details on how our active learning paradigm performs with noisy data, refer to Section \ref{Noisy}. Also, it is important to keep in mind that QBag is a version space method and was not developed with noisy data in mind. Most of the data we will be dealing with is deterministic, except for an occasionally mislabeled point due to numerical errors in the DEM-based simulations.

\color{black}
\section{Proposed Paradigm}
\label{MapGeneration}
\noindent
This section will focus on some of the challenges that were encountered while developing our active learning paradigm. We will discuss several topics that are relevant to the application, such as classifier selection, dealing with noisy labels, and exploration {\em vs} exploitation. This section will conclude with pseudocode for our active learning paradigm.
\label{The Mobility Problem}
\subsection{Model Selection}
\noindent
Selecting the right machine learning algorithm is an essential part of making accurate predictions. For this application, the main challenge is in generating a large enough training set, so the learning algorithm needs to be resourceful with the limited data it has available. Our scheme uses a classifier because many mobility maps use distinct colors to classify speed ranges, like the one shown in Fig. \ref{SimMap}(b). For this type of mobility map, accurately resolving the decision boundaries is far more important than knowing the precise speed throughout the feature space. Therefore, a classifier that can query instances along the decision boundaries would likely require less data than a regression model that might spread itself thin by attempting to make accurate predictions throughout the entire feature space.

In \cite{jayakumar}, authors tested the efficacy of various classifiers to generate mobility maps including kNNs, SVMs, MLPs, kriging, decision trees and random forests. They found that MLP provided the most accurate predictions followed by SVM. In our own tests using these classifiers, we obtained similar results. Only two features---the longitudinal slope and the cone index---were considered in \cite{jayakumar}.  However, additional features are usually required in practice and little is known about the target function in those cases. Therefore, a general purpose classifier is needed. As a result, we ruled out SVM classifiers with linear, quadratic or cubic kernels, since they cannot accurately model disjoint regions of the same class. Instead, we focused on MLP and SVM with a Gaussian kernel. These two classifiers were compared in Section \ref{Results} using preliminary data.

\subsection{Overfitting}
\noindent
Because little is known about the target function in advance, the classifier needs to be complex enough to accurately capture the underlying trends in the data. While on the other hand, if the classifier is too complex, it may ``memorize'' individual instances and overfit the data. This is especially concerning since our training sets will be limited in size. As a result, we utilized several techniques to avoid overfitting while still ensuring that the model can capture more complex patterns in the data.

To begin with, the MLP classifier will use a rectified linear unit (ReLU) as the activation function. Unlike sigmoid and tanh activation functions, ReLU is sparsely activated \cite{nair2010rectified}. This means that neurons can be turned off during the training process which reduces the complexity of the model and helps to prevent overfitting.

Bagging will be used to reduce the importance placed on individual data instances. Bagging uses an ensemble of classifiers to make predictions about new data instances and works by training each classifier in the ensemble on a subset of the training set. Usually these subsets are constructed by randomly selecting instances from the training set with replacement.  To make predictions about new instances, the ensemble takes a majority vote among its members. This averaging effect reduces the variance and helps to prevent overfitting.

Finally, both SVM and MLP classifiers will use a grid search to tune model parameters. While performing the grid search, the accuracy of the models will be measured using a k-fold cross-validation. This approach will allow us to approximate the accuracy of the model without having to use additional resources to construct a validation set. For the SVM, the penalty parameter $C$ will be updated using a grid search each time a new batch of unlabeled instances is chosen. For the MLP, the number of neurons will be viewed as a parameter and will be determined using a grid search as well. By allowing the MLP to start with a few neurons and gradually add more as additional data becomes available, we are able to avoid issues with overfitting while still allowing the model to increase in complexity when needed. Refer to the Section \ref{Results} for more details.

\subsection{Noisy Labels}
\label{Noisy}
\noindent
Noisy data, due to numerical errors or anomalies in the simulations, can have a detrimental effect on the accuracy of any learning algorithm. This is especially the case when using active learning, since the querying techniques are designed to seek out instances that will greatly affect the accuracy of the classifier. Based on data that was obtained in \cite{jayakumar}, we estimated that approximately 2\% of data instances would be mislabeled. While agnostic active learning techniques that are designed for noisy data sets do exist (e.g., \cite{balcan2009agnostic}), they tend to converge more slowly than aggressive techniques, such as QBag, when labels are mostly deterministic. 

As previously mentioned, we used QBag to query instances and bagging to make predictions. Because each classifier in the ensemble only sees a subset of the training set, this reduces the impact a mislabeled instance may have on querying decisions and model predictions. In addition, when mislabeled instances do influence the querying process, they often lead the active learner to select unlabeled instances near the mislabeled point. If the incorrect label was the result of a random error and not a more fundamental issue with the oracle, then labeling instances near a mislabeled point can provide additional evidence for the right class. To make sure our active learning paradigm can perform well when instances are occasionally mislabeled, we tested it using an oracle that incorrectly labeled instances 10\% of the time. The results from these experiments are given in Section \ref{Results}.

\subsection{Batch-Mode Sampling}
\label{Batch}
\noindent
In order to generate enough data for the learning algorithm, we will need to run multiple simulations at a time. However, choosing instances that are close to each other or a labeled point can give redundant information if the label happens to match the class of its neighbors. To avoid this, let us first define the region of disagreement. Let $D$, the region of disagreement, be the set of points in the feature space $X$ where no more than half of the committee members agree on a label. That is,
$$D=\left\{x\in X \left|\max_{y\in Y}\left| \left\{ t\leq n_c|h_t(x)=y\right\}\right|\leq n_c/2\right.\right\},$$
where $t$ is a positive integer, $h_t$ is the $t$th classifier, and $n_c$ is the size of the committee. To avoid redundancies, we will choose instances in the region of disagreement that are both far from labeled instances and other instances that will be queried in the same batch. We do this by first choosing the unlabeled instance in the region of disagreement that is as far as possible from all other labeled instances. Next, we choose the instance that is in the region of disagreement and that is also as far as possible from any labeled instance and the first queried instance. This process repeats until the number of sampled points equals the batch size. We summarize this process in Algorithm \ref{algo2}. To determine the batch size, we ran experiments on the test function shown in Fig. \ref{Test}. Refer to Section \ref{Pretests} for more details.

\begin{algorithm}[ht]
\label{algo2}
{\bf Input:} Number of points to sample: $n$\\
A set of unlabeled instances to choose from: $\mathcal{U}$ \\
A set of labeled instances: $\mathcal{L}$\\
{\bf Initialize} $Q_1=\emptyset$\\
\For{$k=1,\dots,n$}{
$x_k^*=\arg\max\limits_{x_k\in \mathcal{U}} \left(\min\limits_{x\in \mathcal{L}\cup {\mathcal{Q}_k}} \left\|x-x_k\right\|_2\right)$\\
$\mathcal{Q}_{k+1}=\mathcal{Q}_k\cup \{x_k^*\}$
}
{\bf Output:} A set of well-spaced instances: $\mathcal{Q}_{n+1}$
\caption{$\text{MaxMinSample}(n,\mathcal{U},\mathcal{L})$}
\end{algorithm}
\subsection{Exploration vs Exploitation}
\label{Exploration}
\noindent
Sampling bias is an issue with many active learning strategies. By allowing the learning algorithm to query new instances, there's often a good chance that the final training set will not accurately represent the true underlying distribution of the target function. As a result, this can cause the active learner to converge to a suboptimal hypothesis. In practice, this often means that the learning algorithm never discovers that it incorrectly classified a portion of the feature space, because the sampling scheme only queries points outside of the misclassified region. This tends to happen when the active learner queries too many points near the known decision boundaries and ignores more distant points that might expose the learner's oversight.

One way to avoid this is by occasionally querying points outside the region of disagreement. However, this can be tricky since querying too many points will slow down the convergence of our scheme, but not querying enough points may cause the learner to misclassify large portions of the feature space. This conundrum is known as the exploration vs. exploitation dilemma. In this case, exploration is referring to the process of querying over a large portion of the feature space so that all misclassified regions are discovered in a reasonable time. Exploitation, on the other hand, is the process of querying over small regions with promising results, such as the region of disagreement, in order to refine the ensemble's predictions.

To find the right balance, we performed experiments on the test function discussed in Section \ref{prelim}. We found that if we used 1/8th of the queried points for exploration and the remaining points for exploitation, then the convergence was only slightly slower than what we observed when we didn't use any exploratory points at all. Also, this proportion was still large enough so that the scheme could quickly and consistently discover all 5 classes.

When choosing exploratory points, we used the same approach that we used when doing batch sampling. The only difference is that we now choose points outside of the region of disagreement. We used this approach since we wanted to avoid querying exploratory points near labeled points, since doing so would likely provide us with redundant information.
\begin{algorithm}[ht]
\caption{Active Learning Paradigm}
{\bf Input:} Number of trials: $N$\\
Batch size: $n_b$\\
Number of exploratory points: $n_e$\\
Number of committee members: $n_c$\\
A classifier: $A$\\
{\bf Initialize:} $\mathcal{L}_1$, $\mathcal{U}_1$ with random instances\\
\For{$k=1,\dots,N$}{
\begin{enumerate}[label=\arabic*., leftmargin=0.3cm,rightmargin=0.5cm]
\item Use cross validation to find parameters for $A$.\\
\item Randomly sample from $\mathcal{L}_k$ with replacement to obtain subsamples $\mathcal{L}'_1,\dots,\mathcal{L}'_{n_c}$ each of size $m_c$.\\
\item Train $A$ on each subsample to obtain $h_1,\dots,h_{n_c}$.\\
\item Find $\mathcal{D}$ which consists of the points $x\in \mathcal{U}$ that satisfy $\max_{y\in Y}\left| \left\{ t\leq n_c|h_t(x)=y\right\}\right|\leq n_c/2$.\\
\item $\mathcal{Q}_k=\text{MaxMinSample}(n_b-n_e,\mathcal{D},\mathcal{L}_k)$\\
\item $\mathcal{E}_k=\text{MaxMinSample}(n_e,\mathcal{U}\setminus\mathcal{D},\mathcal{L}_k)$\\
\item $X^*=\mathcal{Q}_k\cup\mathcal{E}_k,\quad y^*=\text{Oracle}(X^*)$\\
\item $\mathcal{L}_{k+1}=\mathcal{L}_k\cup\left<X^*,y^*\right>,\quad \mathcal{U}_{k+1}=\mathcal{U}_k\setminus X^*$\\
\end{enumerate}
}
{\bf Output:} $h(x)=\arg\max_{y\in Y}\left| \left\{ t\leq n_c|h_t(x)=y\right\}\right|,$\\
where $h_t$ are hypotheses of the $N$th stage.
\end{algorithm}

\section{Experiments}
\label{Experiments}
\noindent
This section will focus on the experiments that were performed using our active learning scheme. We will begin by describing the setup for these experiments along with additional details on how the active learning paradigm was configured. We will then discuss the results from some of the preliminary tests we performed. We used these preliminary tests to gauge the model's performance before running simulations. After observing satisfactory results from the preliminary tests, we then used physics-based simulations to label instances in both 2- and 3-dimensional features spaces. For the 2-dimensional problem, the adhesion factor and friction coefficient were used as the two features. These features differ from the ones that were considered in \cite{jayakumar} and will serve as a credible challenge to our active learning scheme, since the target function is unknown in advance. Finally, we will include a third feature, the soil density, and will compare our results with random sampling.

\subsection{Experimental Setup}
\label{Setup}
\begin{figure}
\begin{center}
\includegraphics[width=0.45\textwidth]{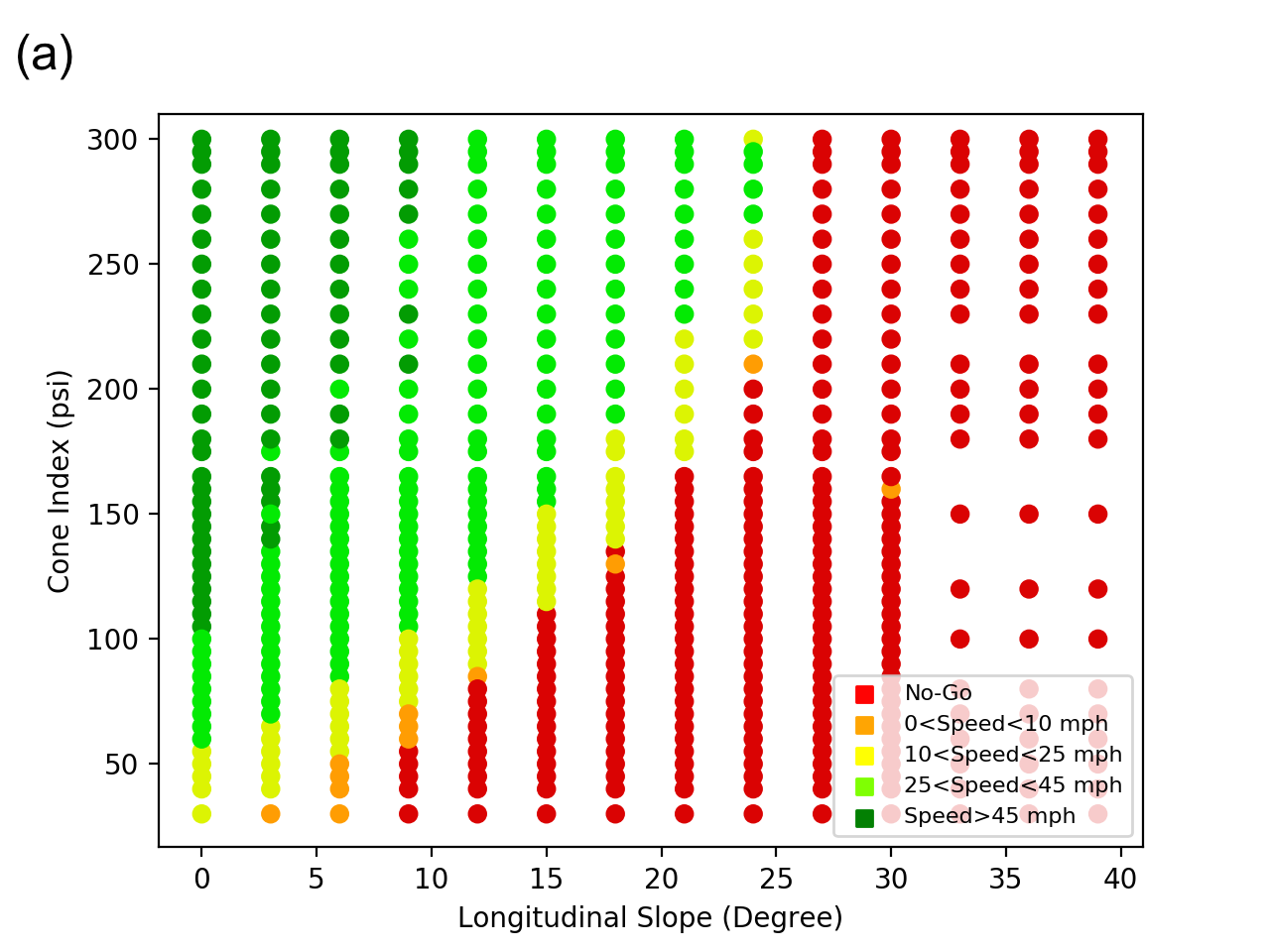}
\includegraphics[width=0.45\textwidth]{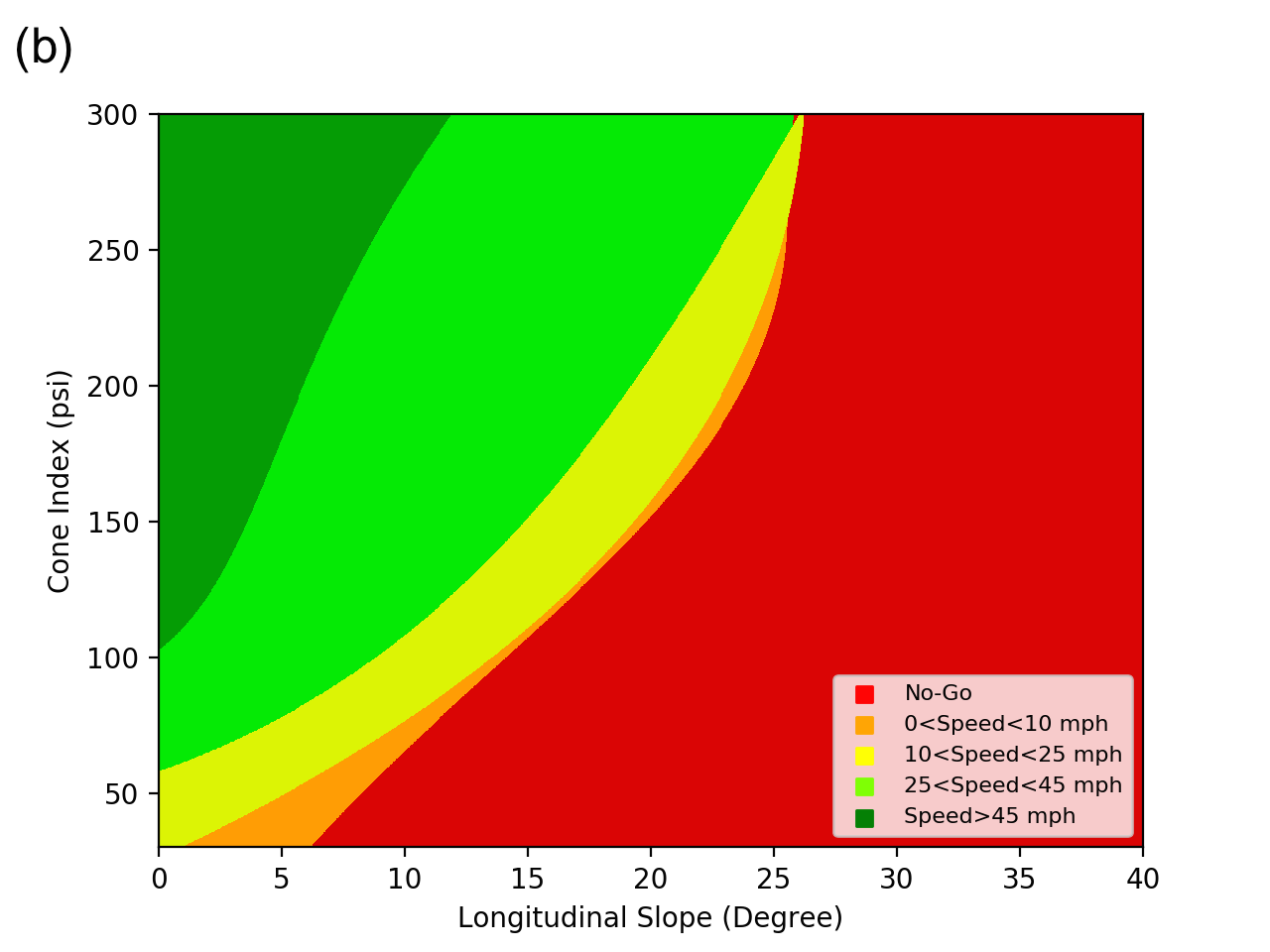}
\end{center}
\caption{(a) This figure shows the original data set that was obtained in \cite{jayakumar}. There is a total of 528 points, and each point was labeled by running a DEM simulation. (b) This figure shows the target function that was used in the preliminary tests. The target function was constructed by training SVM with a cubic kernel on the data set in (a).}
\label{Test}
\end{figure}

\noindent
\subsubsection{Preliminary Tests}\label{prelim} We initially had access to the 528 data points that were generated in \cite{jayakumar}. The points and their labels are shown in Fig. \ref{Test}(a). The labels indicate a range for the speed-made-good based on the longitudinal slope and the cone index. Before running simulations, we constructed a test function in order to gauge the performance of our active learning scheme. The test function was constructed by training SVM with a cubic kernel on the 528-point data set and is shown in Fig. \ref{Test}(b). We felt that this test function would serve as a good predictor of the model's performance, since the original data set was generated using the same DEM-based model.

For the preliminary tests, we used a MLP and a SVM with a Gaussian kernel to predict the speed-made-good. The test function in Fig. \ref{Test}(b) was used as the ground truth. For the MLP classifier, we used a single hidden layer and determined the number of nodes by performing a 10-fold cross-validation on the labeled pool. We started with just 4 nodes in the hidden layer, and with each new batch, we either halved, doubled, or left the number of nodes unchanged, depending on which option gave the lowest cross-validation error. As a safeguard, we never let the number of nodes drop below 2, even though this did not appear to be a significant issue, since the number of nodes tended to gradually increase as more labeled data became available. This approach was adopted because unlike supervised learning where the training set is known in advance, little may be known about the target function and its complexity, so choosing a reasonable number of nodes without eventually overfitting or underfitting the available data can sometimes be challenging. For the 2D test case, we did not see much difference in the prediction accuracy when we used fewer nodes, such as 10, or a larger number, such as 100. However, in a 3D preliminary test case where we extruded the 2D test function, overfitting became more of an issue, and an adaptive approach was needed.

For the SVM, we set the kernel coefficient $\gamma=1/2$ and used a grid search combined with a 10-fold cross-validation on the training data to determine the penalty parameter $C$. The penalty parameter was updated each time a new batch of labeled instances was added to the labeled pool $\mathcal{L}$.

For both classifiers, we set the batch size $n_b=32$, the number of exploratory points $n_e=4$, and the number of committee members $n_c=20$. While smaller committee sizes would likely yield similar results, we were not as concerned with the computational cost of the active learner, since it's cost is negligible in comparison to the simulations. In addition, a large committee can help to average out some of the more extreme predictions made by a minority of its members. This can help to avoid excessive fluctuations in the predictions made by the committee as more training data becomes available. 

To understand how our scheme would perform with occasionally mislabeled data, we ran additional tests where each queried instance was mislabeled $10\%$ of the time. We chose this number, because we felt it was a conservative overestimate of the noise that may result due to numerical errors in the simulations. As a commonly used yard-stick, we compared all of our results to random sampling, where we used the same ensemble of 20 classifiers to make predictions. 

\subsubsection{2D Problem} To test our scheme's performance on a ``real-world'' problem, we ran full length DEM simulations to predict the speed-made-good for a previously unknown target function. This target function depends on two dimensionless variables: the friction coefficient and the adhesion factor. Since we could only perform this test once, due to computational constraints, we used a MLP as our classifier, since it tended to make more accurate predictions with fewer training instances when compared to the SVM. Refer to Section \ref{prelim} for more details.

As mentioned previously, additional tests performed on the target function in Fig. \ref{Test}(b) did not show any significant difference in the accuracy of the predictions when we varied the number of neurons in the hidden layer from 10 to 100. Therefore, we used 100 neurons for the 2D problem in case the target function happened to be fairly complex. It wasn't until later, after the simulations for the 2D problem completed, that we observed that overfitting could pose a more significant challenge in 3D. As a result, the adaptive approach that was mentioned earlier was only utilized for the 3D case and the preliminary tests. However, based on our earlier tests, we do not expect that using 100 neurons in the hidden layer would have any significant effect on the accuracy of our predictions for the 2D case.

Other than the number of neurons in the hidden layer, all hyperparameters will be the same as the ones used in the preliminary tests. This includes the batch size, which was set to 32. We chose this number because it appeared to be a relatively large batch size that consistently gave rapid convergence to the target function in Fig. \ref{Test}. In an ideal setting, a smaller batch size would be preferred, but due to the computational demands involved in labeling points, running simulations in parallel was the only practical option.

\subsubsection{3D Problem} For the 3D case, we added an additional parameter, the soil density, and again used DEM simulations to label queried points. The setup for this test was nearly identical to the 2D case, except for a few things. The first was that we allowed the number of neurons in the hidden layer to vary, depending on the cross-validation errors. Second, we ran many more simulations. To be precise, we ran a total of 1,384 full length DEM simulations. In all, there were 468 simulations that were used to make up the testing set, 448 simulations that were chosen using our sampling scheme, and 448 simulations that were used to construct a randomly generated training set. In addition to that, 20 more simulations were used to construct the initial data set that the QBag and random sampling schemes both started from. Of course, we would expect that more data would be needed due to the curse of dimensionality. However, there may be a slight benefit that comes with this extra dimension. This leads to the third difference, which is the size of the batches. For the 2D case, we used batches of 32. However, for some problems with 3 features, we noticed that quick convergence could often be observed with batch sizes larger than 32. We suspect that this is the case, because the higher dimensionality of the decision boundaries makes it possible for more points to be queried simultaneously without too much redundancy. Because of this and additional tests we performed using basic 3D test functions, we will be using batches of 64 on the 3D case.

\subsection{Results}
\subsubsection{Preliminary Tests}
\label{Results}

\begin{figure}
\begin{center}
\includegraphics[width=0.45\textwidth]{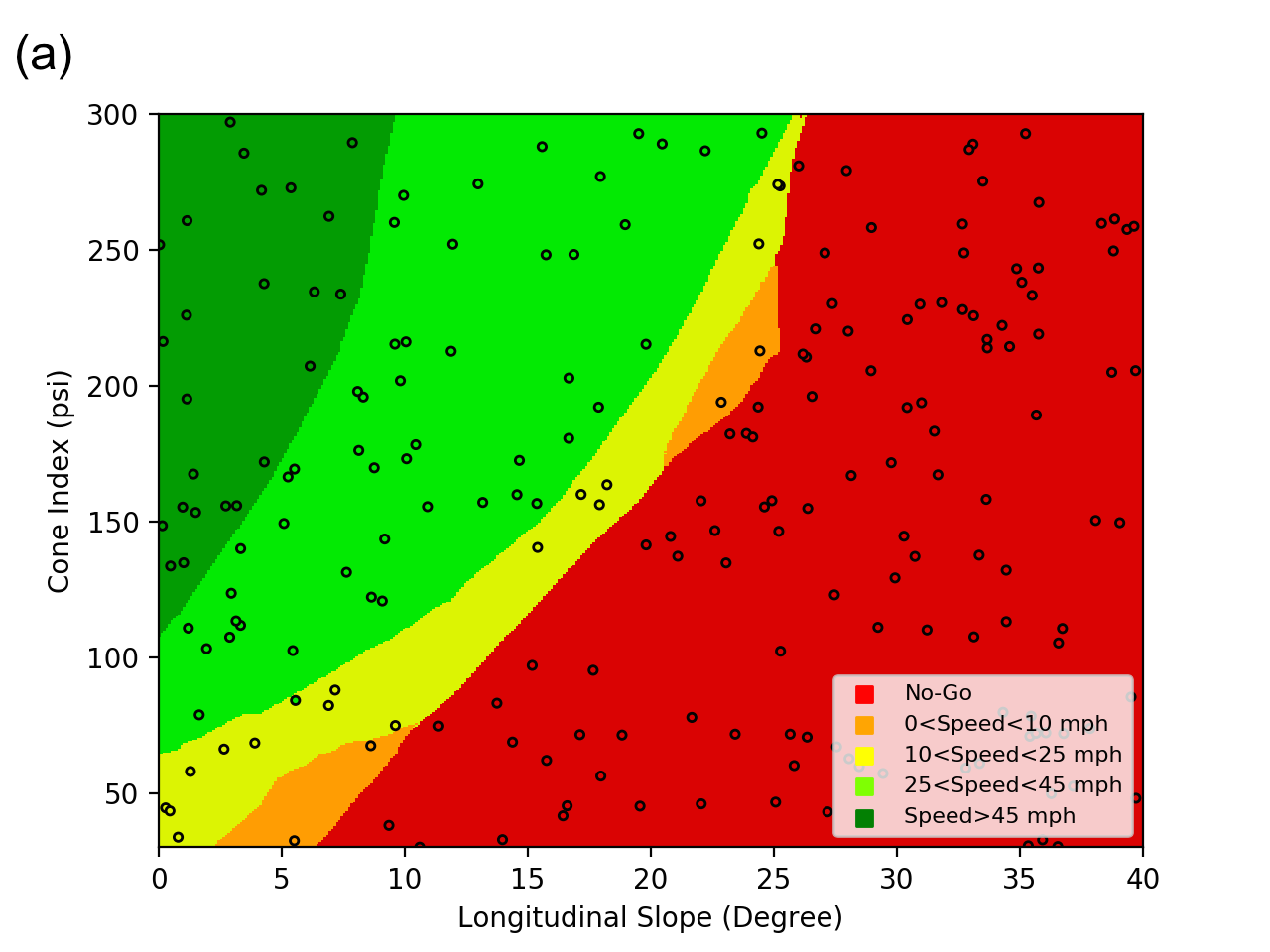}
\includegraphics[width=0.45\textwidth]{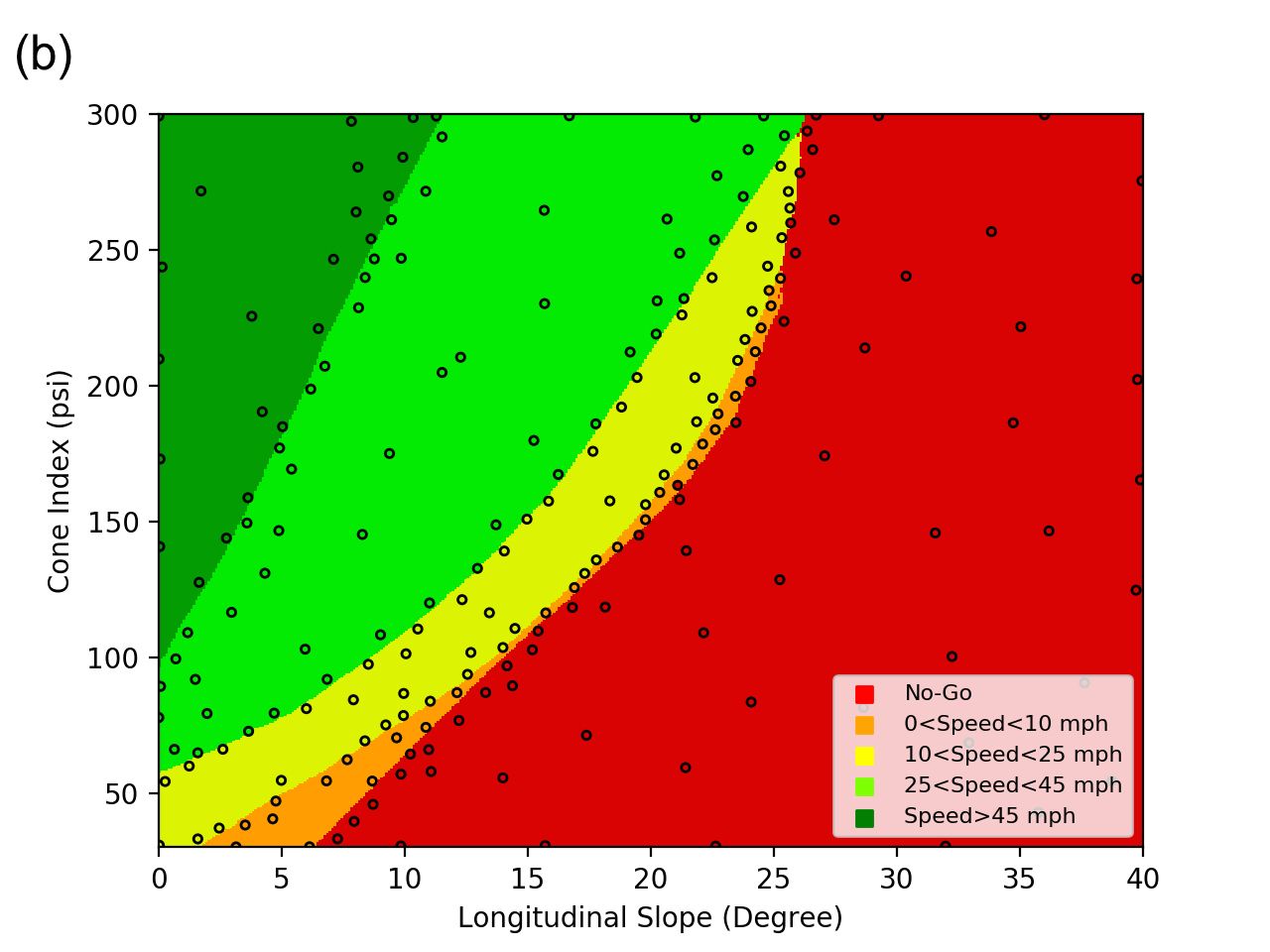}
\end{center}
\caption{These two figures show a comparison between random sampling (a) and our active learning scheme (b). In both cases, the target function in Fig. \ref{Test}(b) was used as the ground truth. For this comparison, all queried points were correctly labeled.}
\label{Pretests}
\end{figure}

Figs. \ref{Pretests}(a) and \ref{Pretests}(b) show a comparison between random sampling and our active learning approach. The predictions for both figures were generated by an ensemble of MLPs. Notice how the active learning-based approach focuses heavily on the decision boundaries while occasionally sampling points throughout the feature space. In Fig. \ref{Pretests}(a), we can see that the same ensemble had difficulties reproducing the test function when we used a training set consisting of randomly queried points. Notice how the ensemble often guesses where the decision boundary is by placing the line approximately halfway between neighboring points with distinct labels. While this may seem like a reasonable thing to do, and it probably is, the lack of data near the decision boundaries results in a classifier that misses many of the details that are captured by our active learning-based approach. As a result, our sampling scheme achieved an accuracy of 99.0\% with only 212 labeled data points, and random sampling achieved an accuracy of 95.7\%. While this may not seem like a huge difference, obtaining that extra 3.3\% can be extremely difficult when the accuracy is so high to begin with.

\begin{figure}
\begin{center}
\includegraphics[width=0.45\textwidth]{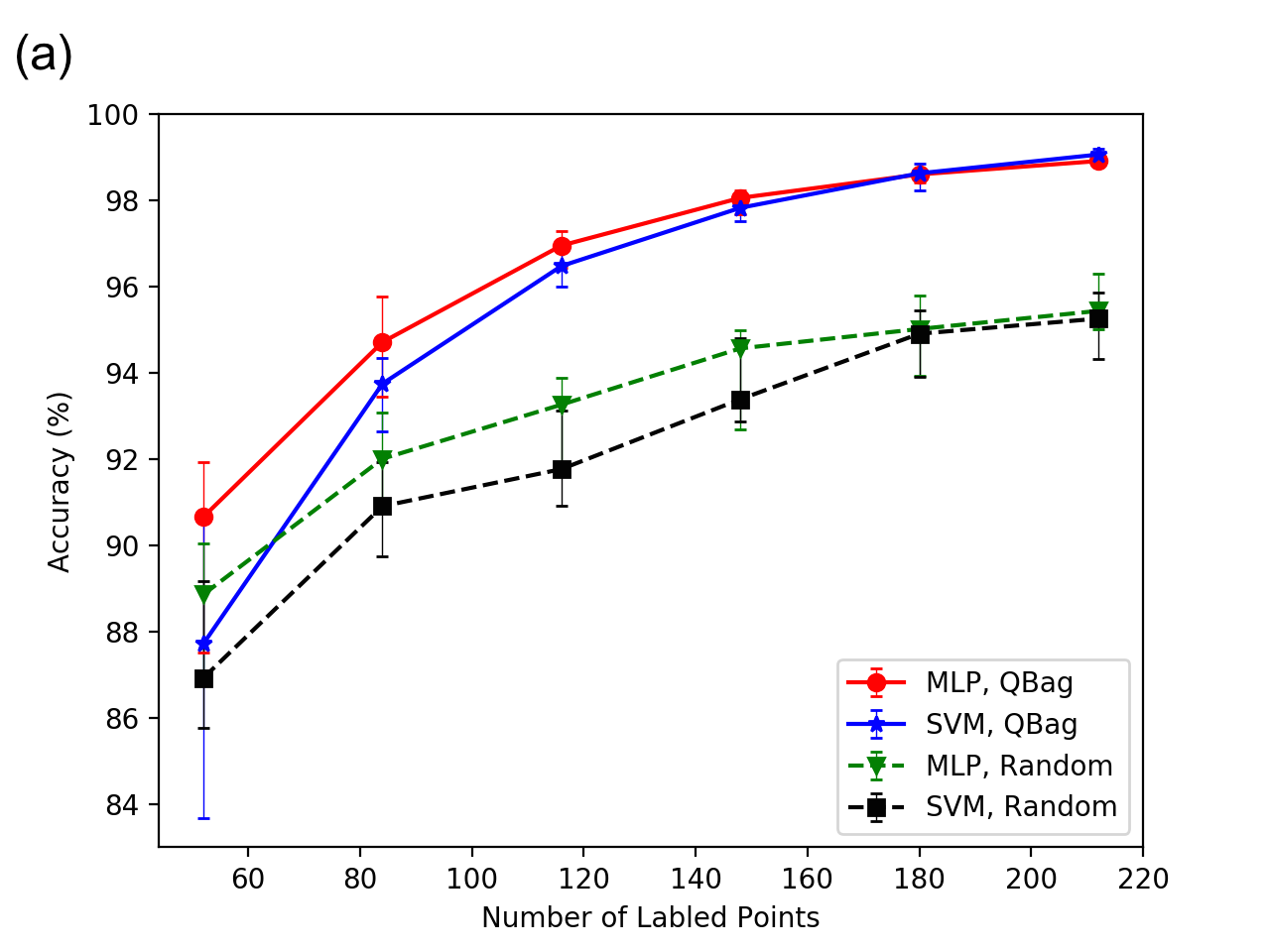}
\includegraphics[width=0.45\textwidth]{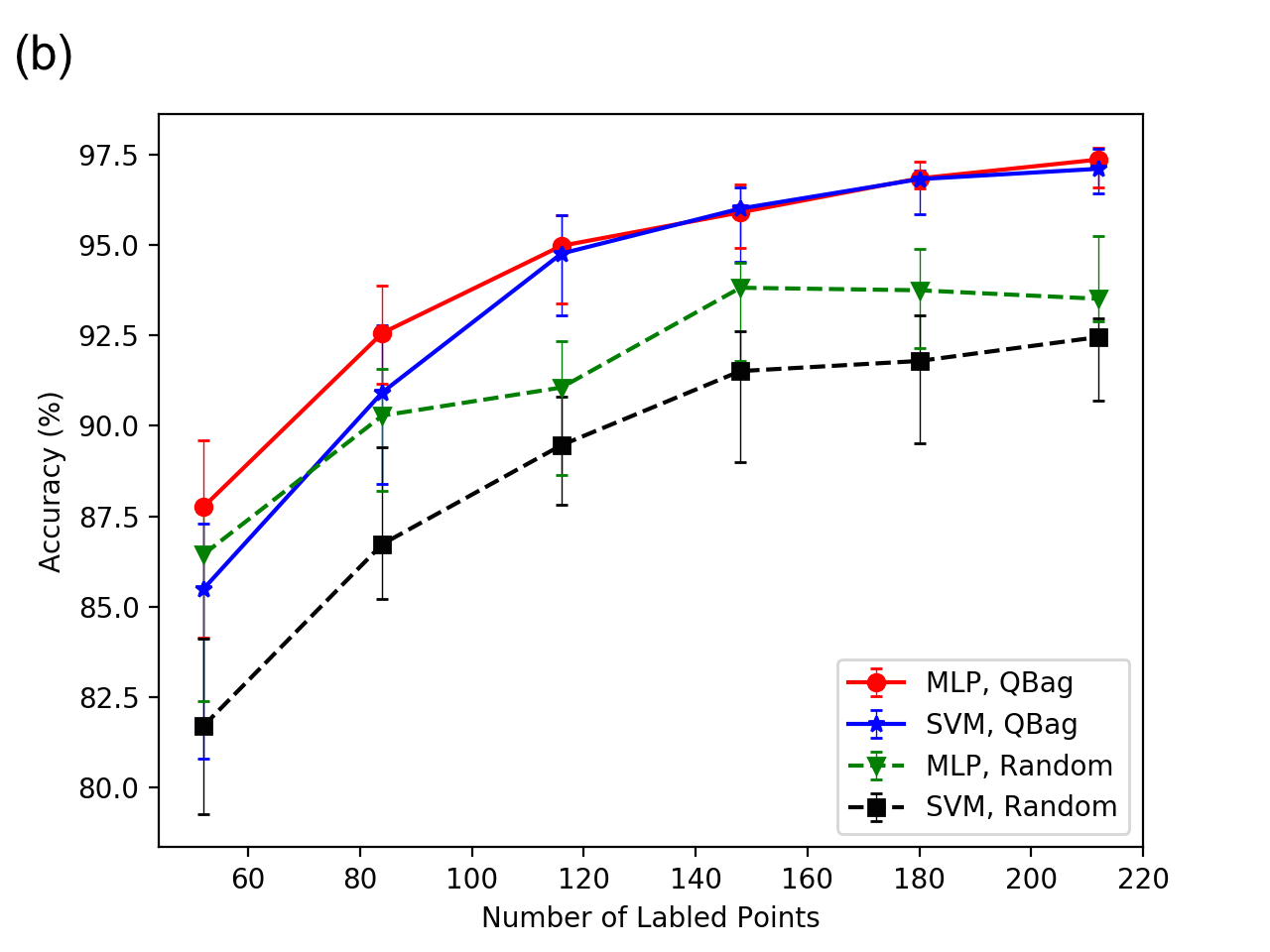}
\end{center}
\caption{These two figures compare the median accuracy of the tested schemes as the number of labeled instances is increased. (a) Shows the accuracy when all queried points are correctly labeled. (b) Shows the accuracy when 10\% of the data is mislabeled. The error bars in both figures show the 50\% confidence interval when the learning algorithms were trained 30 times with distinct, randomly generated initial data. Notice that the improvement in accuracy for the MLP is reduced when some of the training set is mislabeled. Mislabeled data can erode away at some of the gains in accuracy that are achieved through active learning. Therefore, it's imperative that the number of mislabeled instances be kept to a minimum.}
\label{MedAcc}
\end{figure}

Fig. \ref{MedAcc}(a) shows how the number of queried instances impacts the error rates for the ensemble of classifiers. To generate these plots, each ensemble was trained 30 times starting with distinct, randomly generated initial training sets. Each line indicates the median accuracy that was obtained by the corresponding ensemble, and the error bars show the 50\% confidence interval. Notice how both MLP and SVM classifiers attained higher accuracies with lower variation when they were trained using our active learning scheme. After being training on 212 points, the MLP reached a median accuracy of 98.9\% when trained with our scheme, and the SVM performed slightly better with a final accuracy of 99.1\%. With random sampling, the MLP maxed out with a median accuracy of 95.4\% and the SVM attained a median accuracy of 95.3\%. Now suppose that the desired accuracy is set at 95\%. Using random sampling, this is first achieved by the MLP after running 180 simulations and by the SVM after running 212 simulations. With our active learning-based approach, 95\% accuracy is first surpassed using 116 simulations. %At a cost of \$50 per simulation, our active learning scheme would save \$3,200 in computational expenses.

In Fig. \ref{MedAcc}(b), we again compared random sampling with our active learning scheme. However, this time, we trained the classifiers with noisy data. As before, we used the same ensemble of 20 MLPs. The difference is that we intentionally provided the learner with incorrect labels 10\% of the time. This was to simulate noisy labels due to numerical inaccuracies in the DEM-based simulations. We will see later that this in fact became an issue once we started labeling instances using simulations. However, despite being trained with partially incorrect data, our active learning scheme still outperformed random sampling, though in general we wouldn't necessarily expect the difference to be as large as it was in the noise free case. Part of the reason we're seeing such a considerable improvement in performance, despite the noisy data, may be due to the ensemble creating a small region of disagreement around the mislabeled instances. This could coax the learning algorithm into querying additional points nearby. If the mislabeled point was simply the result of a random error, then querying more points nearby could provide additional evidence for the correct label.

Based on the results from Figs. \ref{MedAcc}(a) and \ref{MedAcc}(b), we decided to mainly focus on the MLP classifier in the remaining tests. The reason for this is that the MLP typically produced more accurate results sooner than the SVM. This is an important feature when generating mobility maps, since in most cases it will only be possible to run a limited number of simulations.

\subsubsection{2D Problem}
\begin{figure}
\begin{center}
\includegraphics[height=0.35\textwidth]{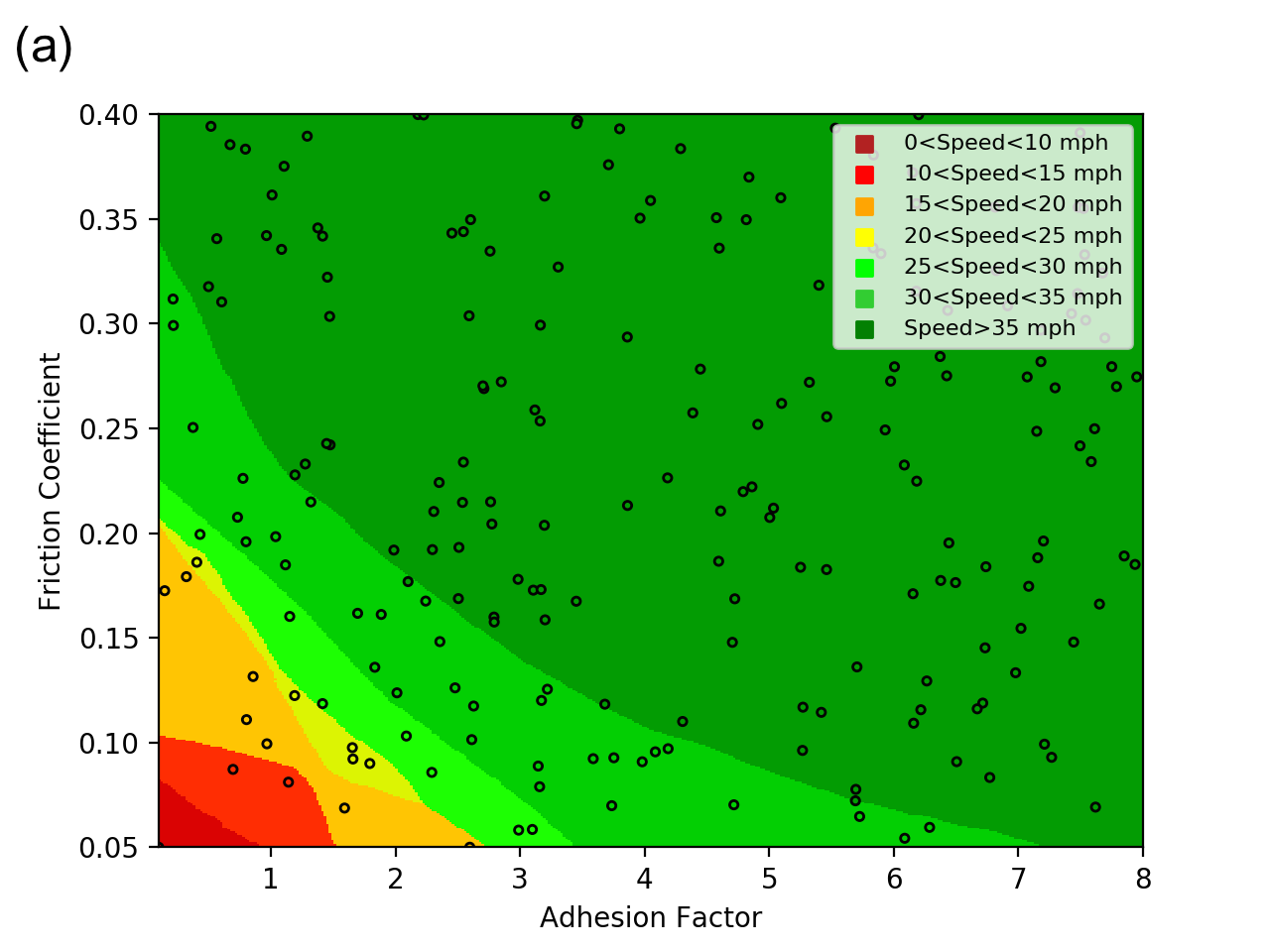}
\includegraphics[height=0.35\textwidth]{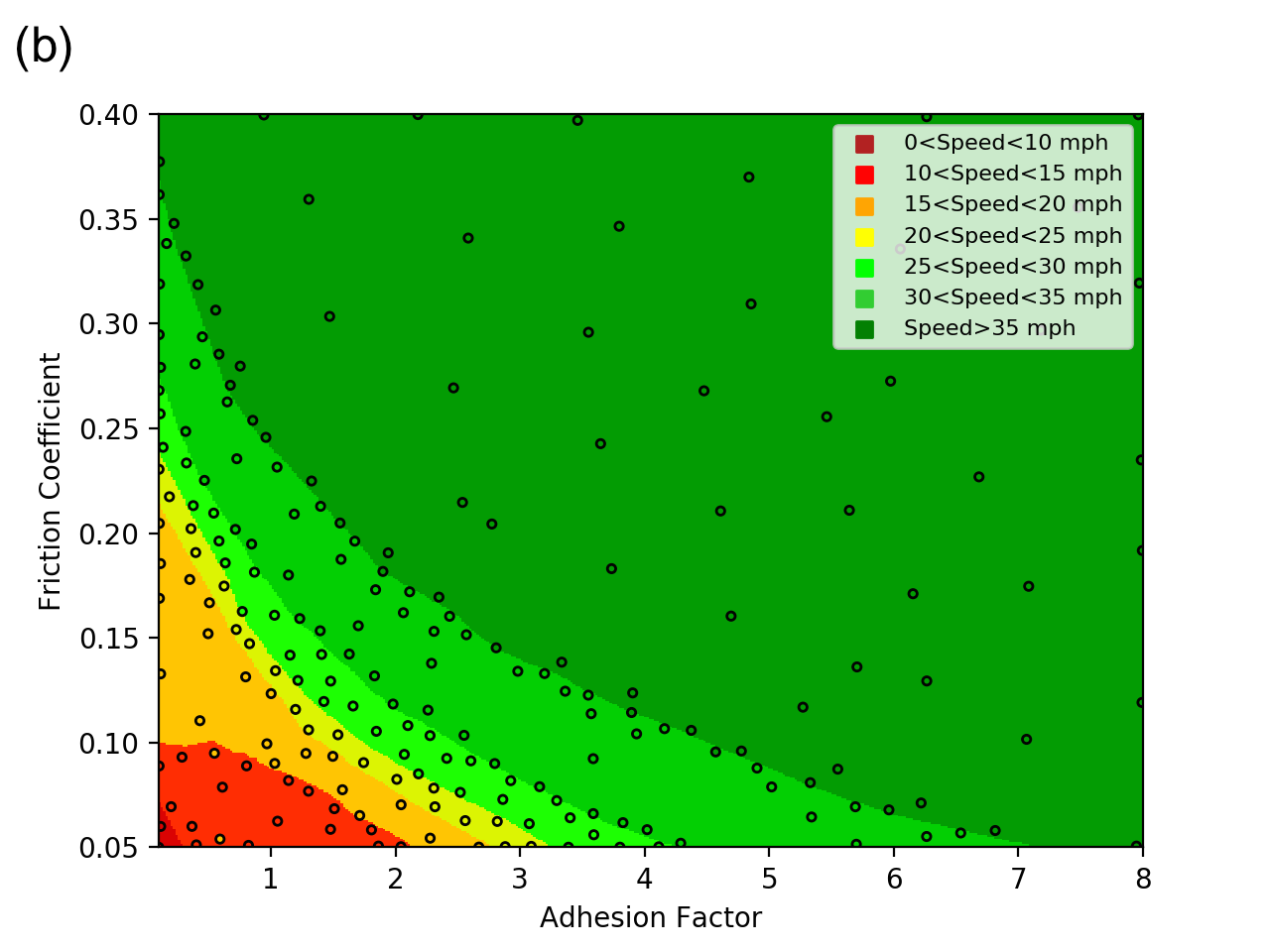}
\end{center}
\caption{These two figures compare the predictions that resulted from using random sampling (a) and our active learning paradigm (b). Each point was labeled by running a DEM simulation. Because simulations could be performed in parallel, we queried points in batches of 32. In each batch, 28 points were selected using active learning and the remaining 4 were chosen outside the region of disagreement using Algorithm \ref{algo2}. Each figure was generated by training an ensemble of MLPs on 212 labeled instances.}
\label{2Dresults}
\end{figure}

In Figs. \ref{2Dresults}(a) and \ref{2Dresults}(b), we compare the predictions made by the MLP ensembles after querying points randomly and with our active learning scheme, respectively. In each case, we trained on the same initial set of 20 randomly sampled points. After that, an additional 192 instances were queried for each approach. To measure the accuracy of the predictions, we constructed a separate testing set that consisted of 212 randomly generated points. We found that the active learning scheme did significantly better than random sampling. To be precise, the active learning scheme achieved 98.1\% accuracy on the testing set, while random sampling only reached 93.4\%. 

\begin{figure}
\begin{center}
\includegraphics[width=0.45\textwidth]{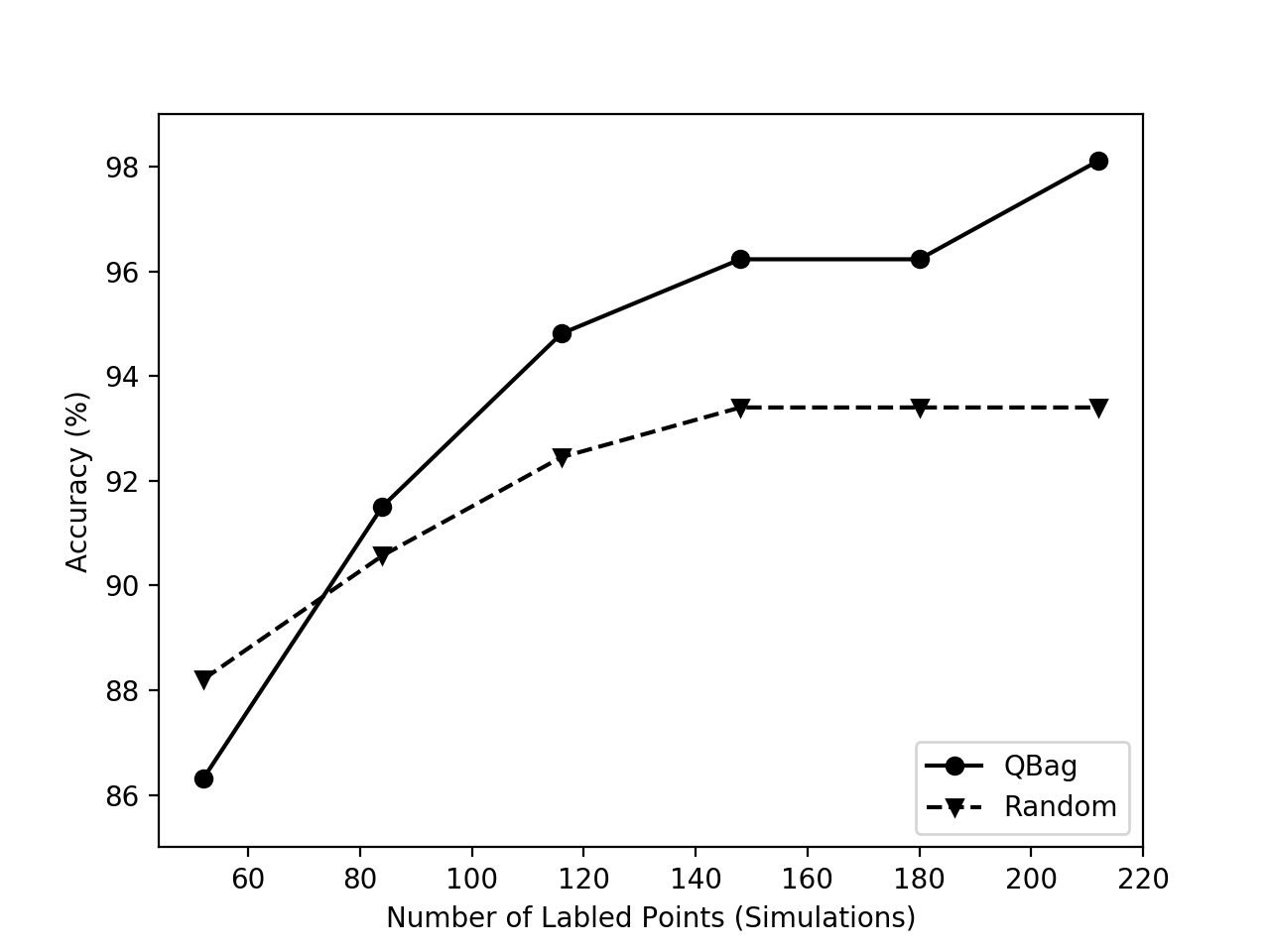}
\end{center}
\caption{This figure compares the convergence behavior of random sampling with our active learning paradigm. Notice that our active learning paradigm obtains a higher accuracy with only 116 points than random sampling with 212 points.}
\label{2DACC}
\end{figure}

The prediction accuracy vs. the number of simulations is shown in Fig. \ref{2DACC}. Notice that after being trained with 116 points, our scheme surpassed the highest accuracies obtained by random sampling. Therefore, to reach 95\% we would need to sample at least 65 more points with the random scheme than we would with the active learning approach. 

Finally, if we refer back to Fig. \ref{2Dresults}, there's one more thing to discuss. Near the point $(0.5,0.05)$, there appears to be a data point that was mislabeled. (The point near $(0.5,0.09)$ may have been mislabeled as well.) As mentioned earlier, the potential for noisy data was a concern from the beginning. In this case, it appears that the mislabeled point may be the result of the time step being too large in the DEM simulations, since the simulations tend to lose stability when the friction coefficient and adhesion factor are both small. This could potentially be addressed by applying a voting filter to the data in order to identify points with questionable labels \cite{votefilter}. The speed-made-good could then be reevaluated using a smaller time-step.

\subsubsection{3D Problem}
\begin{figure}
\begin{center}
\includegraphics[height=0.35\textwidth]{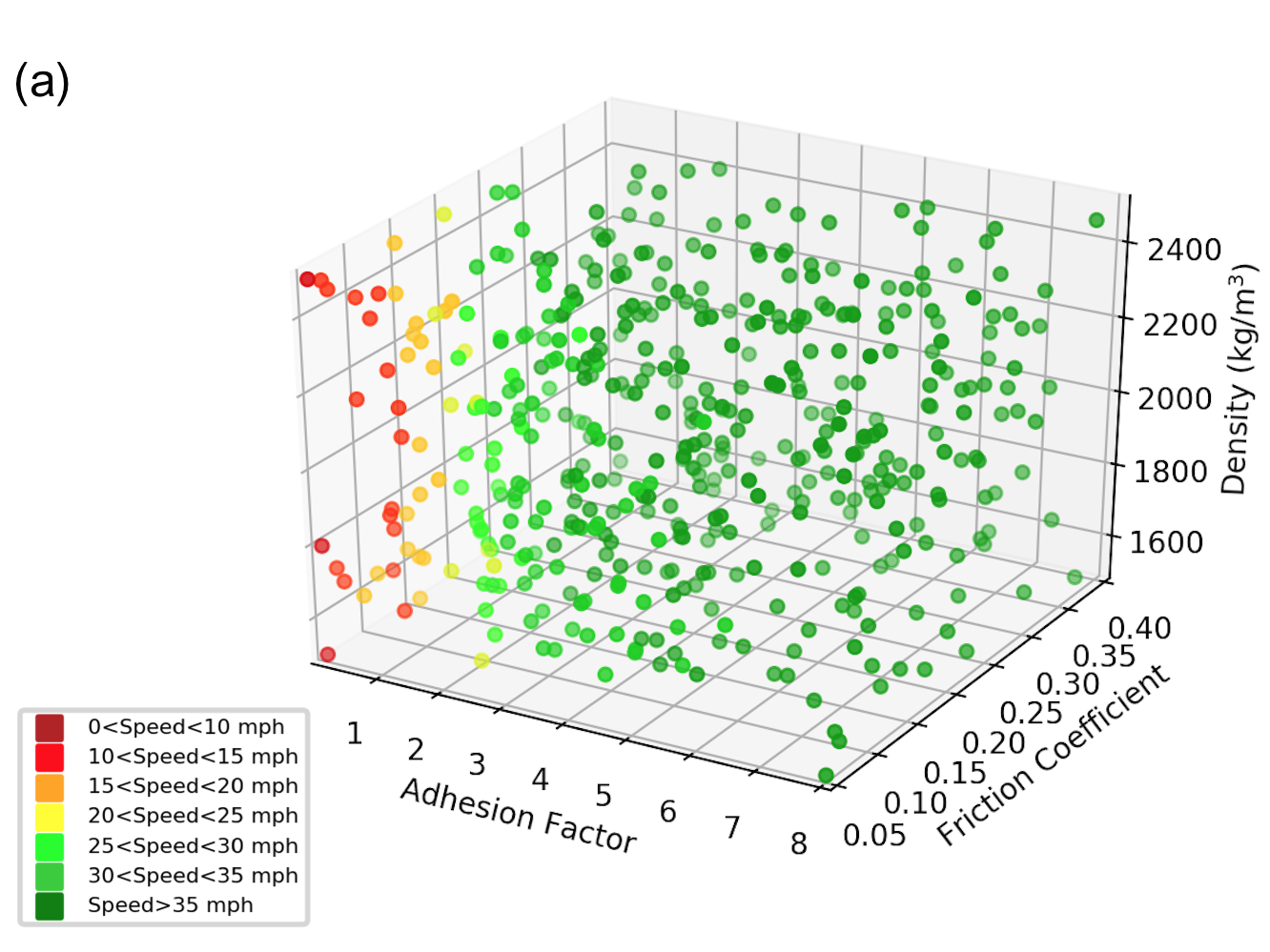}
\includegraphics[height=0.35\textwidth]{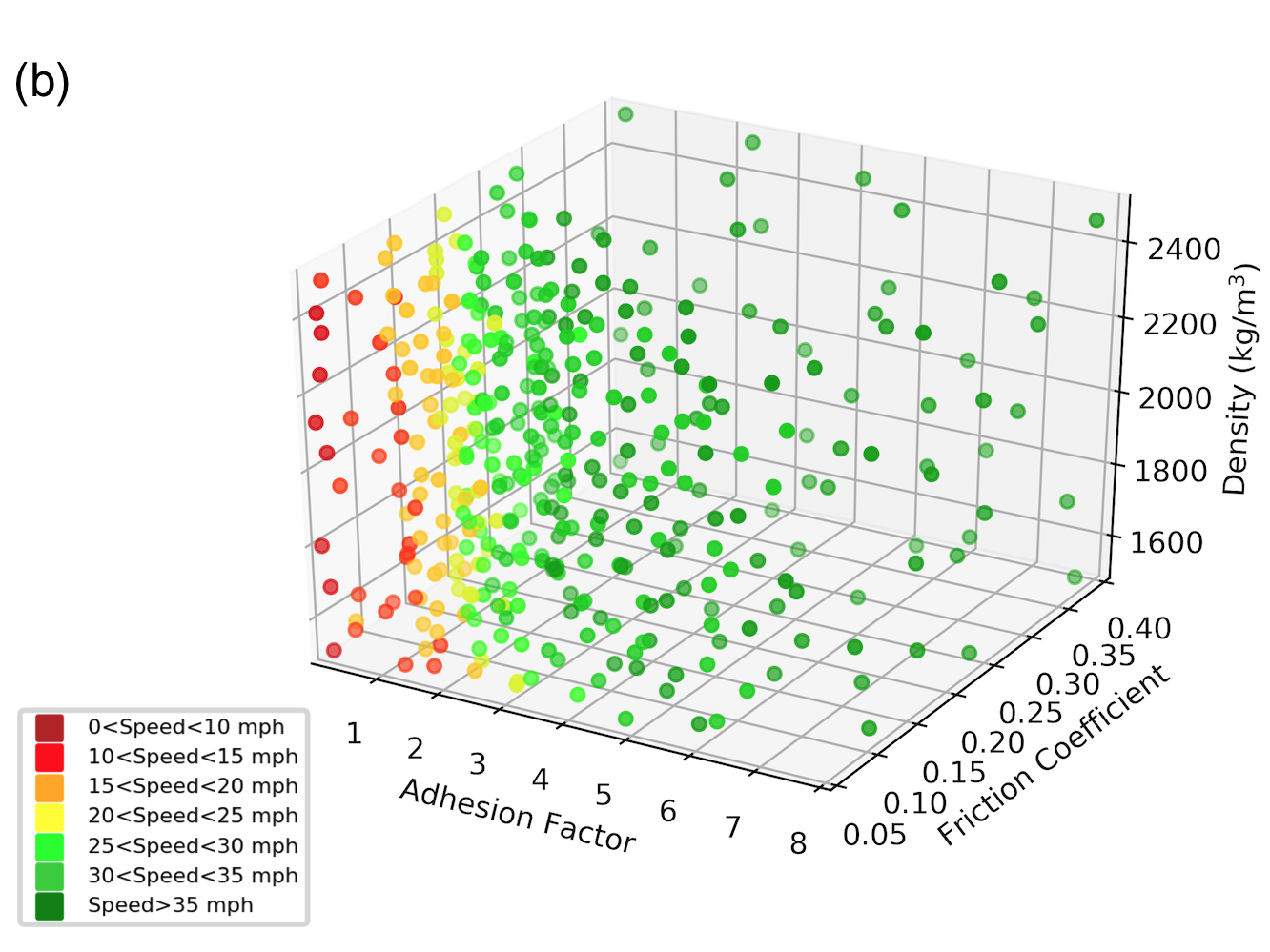}
\end{center}
\caption{(a) A scatter plot of the points that were queried using random sampling. The colors indicate the label that was obtained by running a DEM simulation for each point. (b) A scatter plot of the points that were queried using our active learning scheme. Notice that the scheme tends to query points near the decision boundaries.}
\label{3DScatter}
\end{figure}

In Figs. \ref{3DScatter}(a) and \ref{3DScatter}(b), we plotted the labeled instances that were selected using random sampling and our active learning scheme, respectively. Notice that most of the points selected by our active learning scheme tend to cluster around the decision boundaries. In addition, there are a few instances farther away that serve as exploratory points. For this test, each batch of 64 points had 56 points that were queried within the region of disagreement and 8 points were used for exploratory purposes. 

\begin{figure}
\begin{center}
\includegraphics[width=0.45\textwidth]{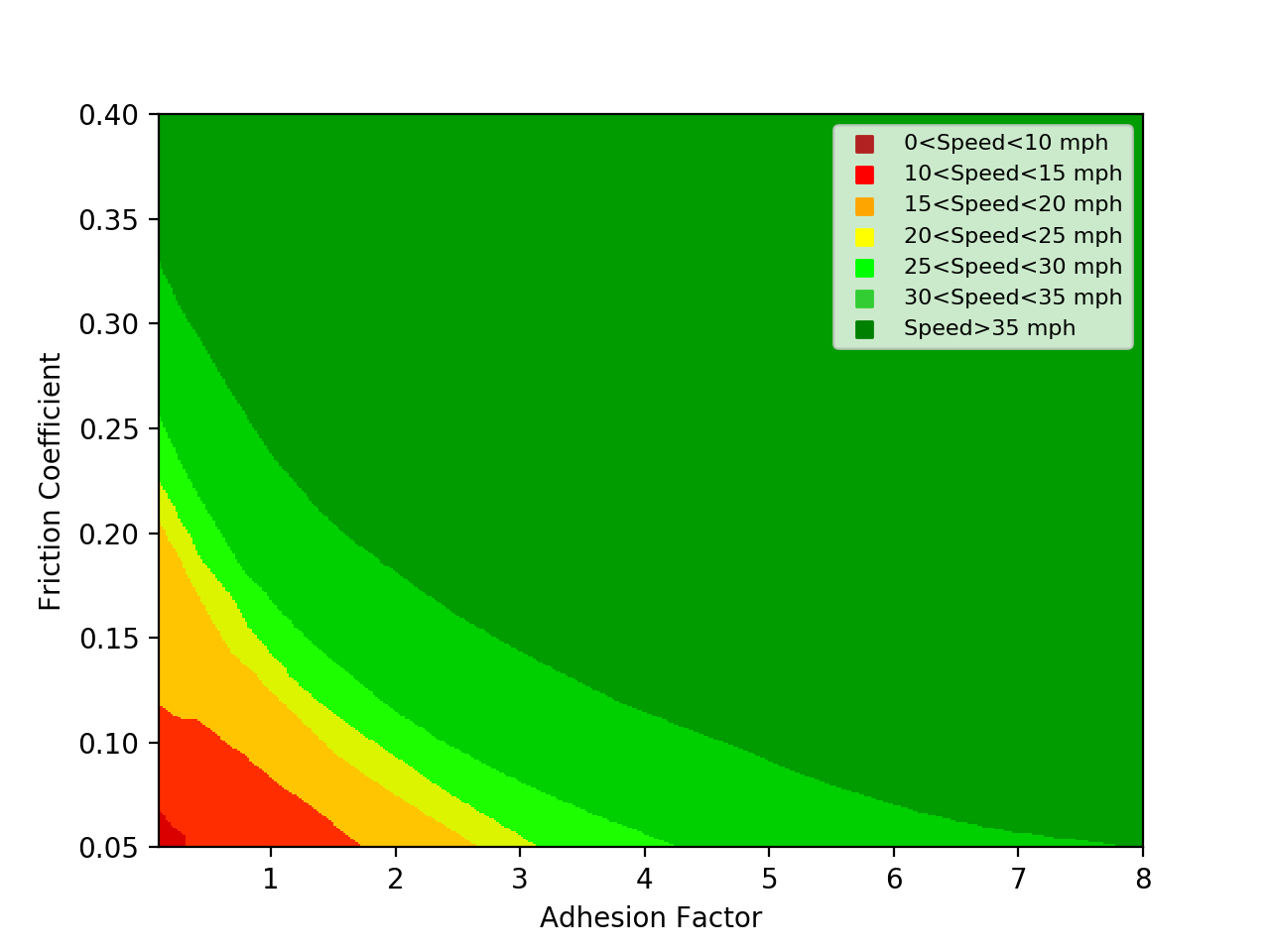}
\end{center}
\caption{This figure shows the prediction that was generated by the 3D model after it was trained using our active learning scheme. In this figure, the density of the soil was fixed at 1800 kg/m$^3$, which corresponds to the density that was used in the 2D case.}
\label{Top}
\end{figure}

Fig. \ref{Top} shows a horizontal slice of the prediction that was generated using our active learning scheme. The figure shows how the friction coefficient and the adhesion factors affect the speed-made-good when the density of the soil is fixed at 1,800 kg/m$^3$. It turns out that in the 2D case, we also fixed the density at 1,800 kg/m$^3$, so this slice should correspond to the prediction shown in Fig. \ref{2Dresults}(b). On first observations, it appears that the main difference is along the decision boundary between the red and orange classes. As mentioned before, this is likely due to inaccurate labels that were the result of numerical instabilities in the DEM simulations. In future tests, it may be beneficial to either reduce the time step when points are queried in that region or to rerun a simulation using a smaller time step when a voting filter considers the label to be questionable.

\begin{figure}
\begin{center}
\includegraphics[width=0.45\textwidth]{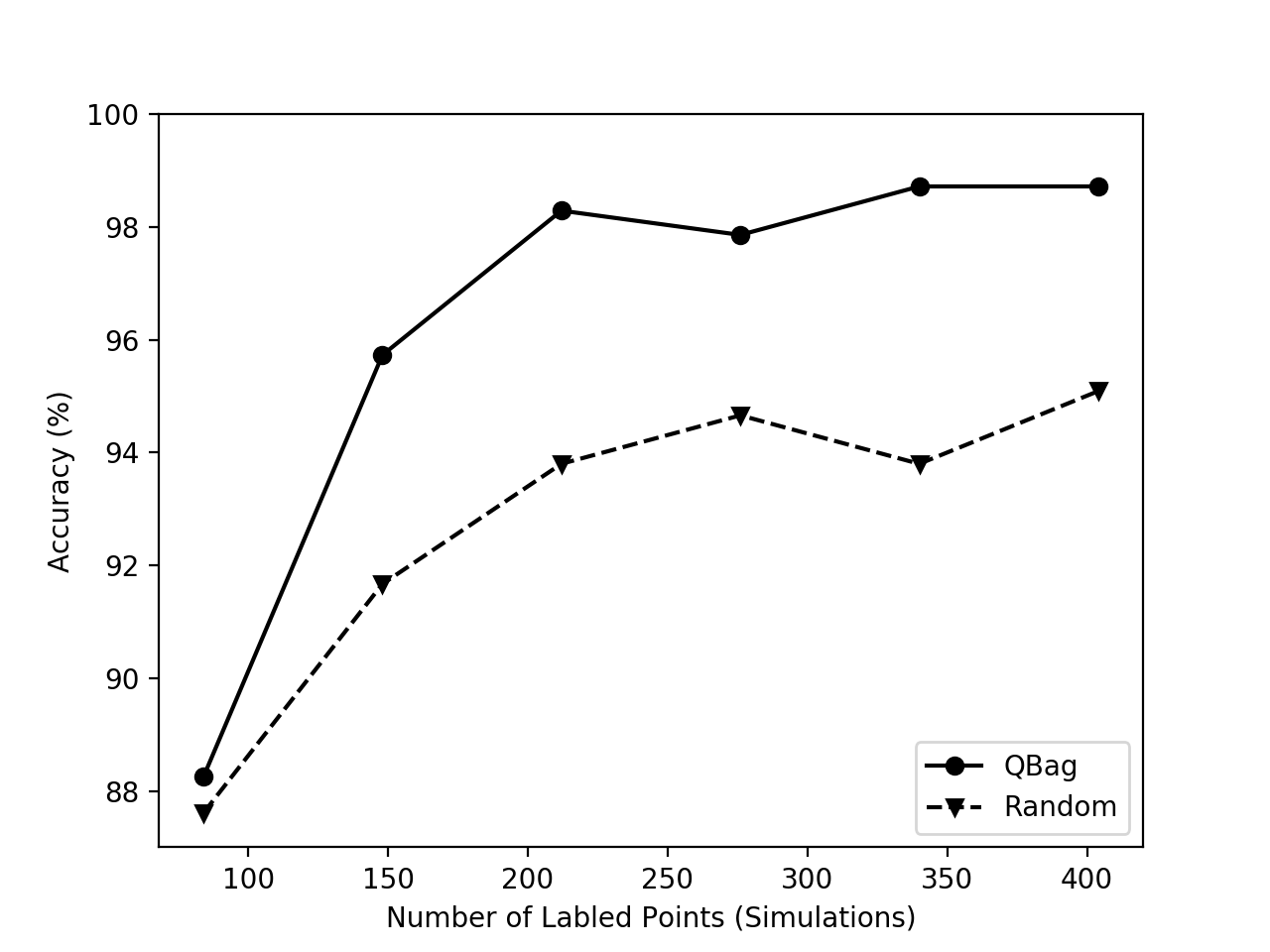}
\end{center}
\caption{This figure compares the convergence behavior of random sampling with our active learning paradigm. Notice that our active learning paradigm exceeds 95\% accuracy with only 148 points. This contrasts with random sampling, which needs 404 points to reach 95\% accuracy. 
}
\label{3DACC}
\end{figure}

Finally, in Fig. \ref{3DACC} we compared the convergence behavior of random sampling with our active learning scheme. We found that our scheme provided a significant benefit over random sampling. As an example, if we used random sampling to train the ensemble with 95\% accuracy, we would need to run 404 simulations to generate the data. On the other hand, if we used our active learning scheme, we would only needed 148 labeled points in order to exceed that same accuracy. That's nearly a reduction in the number of simulations by a factor of 3. 

\label{3Dresults}
\section{Conclusions and Future Work}
\noindent
We have demonstrated that query-by-bagging can be used to significantly reduce the number of physics-based simulations that are needed to construct a mobility map when compared to random sampling. In addition, we have expanded the feature space and are able to accurately predict the speed-made-good using the friction coefficient, the adhesion factor, and the density. Finally, we have provided a framework for generating mobility maps that can be used to incorporate additional soil parameters.

There are a number of interesting directions for future work. To begin with, more work needs to be done to determine which parameters are most important for predicting the speed-made-good. That is, feature reduction or feature extraction techniques need to be utilized to reduce the dimension of the feature space. This could be done using a technique such as principal component analysis or by training an autoencoder.

Another direction could focus on expanding the feature space to include multiple vehicle designs or multiple soil types (mud, snow, sand, etc.). This could be done by using a one-hot encoding. By using a single classifier with multiple vehicle designs or soil types, it may be possible to reduce the overall number of simulations. This is because some of the information that's learned for a single vehicle design or soil type could prove to be useful when training the learner about a new vehicle or soil type.

Finally, it may be possible to use transient data to predict the speed-made-good. When running simulations, we made sure that the vehicle's speed came to a steady-state before labeling the point. However, there were many cases where the final label was obvious long before the vehicle reached a steady-state. Therefore, it may be reasonable to train a classifier to predict the label for the speed-made-good long before the simulation reaches a steady-state. This could help to significantly reduce the runtime for some of the simulations.
\label{Conclusions}
\section{Acknowledgments}
We thank Dave Mechergui for several discussions pertaining to this work and Tamer Wasfy for help with DEM simulations. We acknowledge support from the Automotive Research Center (ARC) in
accordance with Cooperative Agreement W56HZV-19-2-0001 with U.S. Army CCDC Ground Vehicle Systems Center. This research was supported in part through computational resources and services provided by Advanced Research Computing Center at the University of Michigan, Ann Arbor.
\bibliography{activebib}
\bibliographystyle{ieeetr}

\end{document}